\documentclass{article}
\usepackage{spconf,amsmath,graphicx}
\usepackage{url}
\usepackage{booktabs}
\usepackage{amssymb}
\usepackage{color}
\usepackage{soul}
\usepackage[normalem]{ ulem }
 \usepackage{lipsum}% http://ctan.org/pkg/lipsum
\usepackage{multicol}
 \usepackage[table]{xcolor}
\usepackage[utf8]{inputenc}
\usepackage{multirow}
 \usepackage{hyperref}

\title{Predicting ASR Performance on Unseen Broadcast Programs}

\title{ASR Performance Prediction on Unseen Broadcast Programs using Convolutional Neural Networks}
\name{Zied Elloumi $^1$ $^2$, Laurent Besacier$^2$,  Olivier Galibert   $^1$, Juliette Kahn$^1$, Benjamin Lecouteux$^2$  }
\address{
  $^1$ Laboratoire national de métrologie et d\textquoteright essais (LNE) , France \\
 $^2$ Univ. Grenoble Alpes, CNRS, Grenoble INP, LIG, F-38000 Grenoble, France
\\  zied.elloumi@lne.fr}

\begin{document}
%\ninept
%
\maketitle
\begin{abstract}
In this paper, we address a relatively new task: prediction of ASR performance on unseen broadcast programs. We first propose an heterogenous French corpus dedicated to this  task. Two prediction approaches are compared: a state-of-the-art performance prediction based on regression (engineered features) and a new strategy based on convolutional neural networks (learnt features). We particularly focus on the combination of both textual (ASR transcription) and signal inputs. While the joint use of textual and signal features did not work for the regression baseline, the combination of inputs for CNNs leads to the best WER prediction performance. We also show that our CNN prediction remarkably predicts the WER distribution on a collection of speech recordings.
% results. The best performance is obtained with CNN${Softmax}$ (EMBED + RAW-SIG) which outperforms a strong regression baseline (MAE is reduced from 21.99\% to 19.24\%, while $\tau$ is improved from 45.82\% to 46.83\%).
%The task is seen either as a classification problem (predicting range of the WER) or as a prediction problem (predicting the WER itself).
%Both approaches are evaluated using different metrics. The results show that performance prediction on unseen programs is a difficult task, especially on shows containing a spontaneous type of speech. We believe, however, that the provided framework (corpus, metrics) is a good starting point to stimulate research on this new topic.
\end{abstract}
\section{Introduction}
\label{sec:intro}

Predicting automatic speech recognition (ASR) performance on unseen speech recordings is an important Grail of speech research. From a research point of view, such a task helps understanding automatic (but also human) transcription performance variation and its conditioning factors. From a technical point of view, predicting the ASR difficulty is useful in applicative workflows where transcription systems have to be quickly built (or adapted) to new document types (predicting learning curves, estimating the amount of adaptation data needed to reach an acceptable performance, etc.). 

ASR performance prediction from unseen documents differs from confidence estimation (CE). While CE systems allow detecting correct parts as well as errors in an ASR output, they are generally trained for a particular system and for known document types. On the other hand, performance prediction focuses on unseen document types and the diagnostic may be provided at broader granularity, at document level for instance\footnote{Nevertheless this paper will analyze performance prediction at different granularities, from fine to broad grain: utterance, collection.}. Moreover, in performance prediction, we may not have access to the ASR system investigated (no lattices nor N-best hypotheses, no internals of the ASR decoding) which will be considered as a black-box in this study. 

 \textbf{Contribution}
This paper proposes to investigate a new task: prediction of ASR performance on unseen broadcast programs. Our first contribution is methodological: we gather a large and heterogenous French corpus (containing \textit{non spontaneous} and \textit{spontaneous} speech) dedicated to this task and propose an evaluation protocol. Our second contribution is an objective comparison between a state-of-the-art performance prediction based on regression (engineered features) and a new strategy based on convolutional neural networks (learnt features). Several approaches to encode the speech signal are investigated and it is shown that both (textual) transcription and signal encoded in a CNN lead to the best performance.
%Since neural networks are used for text classification, we have proposed a method for converting classes into continuous values (WER) using the probabilities of the Softmax layer of the CNN.
%The idea is to compare  two approaches, one based on regression using the \textit{TranscRater} (\textit{engineered features}) and another based on neural networks using CNN Sentence (learnt  features). \\

\textbf{Outline} The paper is organized as follows. Section~\ref{srw} is a brief overview of related works. Section~\ref{sep} details our evaluation protocol (methodology, dataset, metrics). Section~\ref{sapp} presents our ASR performance prediction methods and~\ref{sear} the experimental results. Finally  section~\ref{sc} concludes this work.

\section{Related works}
\label{srw}
Several  works tried to propose effective confidence measures to detect errors in 
ASR outputs. Confidence measures were introduced for OOV
detection by \cite{Asadi1990} and extended by
\cite{Young1994} 
who used
word posterior probability (WPP) as confidence measure for ASR. 
%Posterior probability of a word is most of the time computed using the hypothesis word graph \cite{Kemp1997}. Also, 
While most approaches for 
confidence measure estimation use side-information extracted from the recognizer  \cite{Lecouteux2009},
%normalized likelihoods (WPP), the number of competitors at the end of a word (hypothesis density), 
%decoding process behavior, linguistic features, acoustic features (acoustic stability, 
%duration features) and semantic features.
%Now, if we focus on methods that do not depend on the internals of the ASR system, we can cite 
methods that do not depend on the knowledge of the ASR system internals were also introduced
\cite{negri2014quality}.% who
% proposes a supervised regression  approach without needing to use the reference transcripts and which does not necessarily depend on the information about the ASR system, this approach has been tested in different scenarios with different difficulties.
%: easy prediction (information from the ASR system are difficult to predict), difficult prediction (homogeneous data and ASR informations are not accessible), difficult prediction on non-homogeneous data, and system informations are not accessible. The data used are the English data provided by IWSLT 2012 and IWSLT2013.
%In order to implement this approach, \cite{negri2014quality} compared the two learning algorithms Support Vector Machines (SVMs) (Shawe-Taylor and Cristianini, 2004) and Extremely Randomized Trees.
 %To characterize utterances, \cite{negri2014quality} 
 %proposed several types of features (using notably ASR output features and signal features). 
 %ASR, Signal, Hybrid, textual. Experiments show that the combination of features produces the best prediction over the use of a single type of features.The results also show that the performance of the prediction models depends on the degree of homogeneity between the training corpus and the test corpus.
%  An analysis of the relation between prediction performance and the size of the train corpus was performed.The results show that the best results were obtained with the 40\% of the corpus of train compared to the best MAE.

As far as the WER prediction is concerned,  \cite{jalalvand2016transcrater} proposed an open-source tool named \textit{TranscRater} based on feature extraction (lexical, syntactic, signal and language model features) and regression. 
% regression (WER prediction) or classification (if multiple ASR outputs are provided). The authors propose four types of features, three of Textual type and one of Signal type.% four types of POS-part-of-speech Features, Module Feature (LM), Lexical Features (LEX) and Signal Features (SIG). %In order to evaluate the performance of the two approaches, the authors use the English data CHiME-3 data.
Evaluation was performed 
%in terms of MAE for the regression approach and in terms of normalized discounted cumulative gain (NDCG) for the classification task. The results show that textual features obtain the best perfomance in both approaches and that the addition of the signal degrades the prediction quality in both regression and classification approaches using 
on CHiME-3 data and interestingly it was shown that \textit{signal} features did not help the WER prediction.
%%%%%%%%%%%%%%% INTERSPEECH %%%%%%%%%%%%%%%%%%%%%%%%%%%%%%%
% \color{red} Finally, \cite{jalalvand2015stacked} proposed a Stacked Auto-Encoder (SAE) and compared it to different types of classifiers (Support Vector Machines, MaxEnt, and Extremely Randomized Tree classifiers) for the WER prediction. 
%The results demonstrated that SAE was better than the other error detection approaches (evaluation made on the ASR outputs of TED talks corpus).
%%%%%%%%%%%%%%%%%%%%%%%%%%%%%%%%%%%%%%%%%%%%%%%%%%%%%%%%%%%%

%\cite{specia2013quest} proposes the \textit{QUEST} tool which is based on the regression approach. It offers different types of Features. The number of features varies from 80 to 123 depending on the case of black-box and go from 39 to 48 depending on the SMT system when the SMT confidence information is available (Glass-box Feature). Feature extraction is carried out at the word level but it can also be in sentence level by calculating the average. In their experiments, the authors use the data WMT12, EAMT11, EAMT09 for the language pair English-Spanish and GALE11 for Arabic-English. The results show that the Gaussian Process technique is promising because it reduces the training time of the model while obtaining similar results compared to models with much more features.

One contribution of our paper is to encode signal information in a CNN for WER prediction. Encoding signal in a CNN has been done in several speech processing front-ends 
%Signal features such as log mel-filterbank energies or mel-frequency cepstral coefficients (MFCCs) are used as a front-end in most speech processing tasks~
\cite{piczak2015environmental,sainath2015learning,jin2016lid}.  Some recent works directly used the raw signal for speech recognition  \cite{sainath2015learning,palaz2015convolutional} or for  sound classification \cite{CNNrawWav}.
% \cite{sainath2015learning} showed that a \textit{CLDNN} on the raw signal and a \textit{CLDNN} using log-mel features have similar performance in terms of WER on both clean and noisy data. \cite{palaz2015convolutional} also demonstrated that a CNN which takes as input the raw signal is better than an ANN which takes as input the MFCCs. \cite{CNNrawWav} proposed a different deep CNN architecture for environmental classification which taking the raw signal as input. Its best model, \textit {m18}, shows a good performance in terms of accuracy (71.68 \%).

\section{Evaluation Framework}
\label{sep}

%\subsection{ Overview }
%\label{sub:Overview}

%\begin{figure}[htb]
%  \centering
%  \includegraphics[width=\linewidth]{img/figure.pdf}
%  \caption{Overview of the Performance Prediction Process}
%  \label{fig:diagram_pred}
%\end{figure}

We focus on ASR performance prediction on unseen speech data. Our hypothesis is that performance prediction systems should only use the ASR transcripts (and the signal) as input in order to predict the corresponding transcription quality. Obviously, reference (human) transcriptions are only available during training of the prediction system. 
%Figure~\ref{fig:diagram_pred} describes the general process:
A \textit{Train$_{pred}$} corpus contains many pairs \textit{\{ASR output, Performance\}} (more than 75k ASR turns in this work), a \textit{Test$_{pred}$} corpus only contains ASR outputs (more than 6.8k turns in this work) and we try to predict the associated transcription performance.  Reference (human) transcriptions on \textit{Test$_{pred}$} are used to evaluate the quality of the prediction.

\subsection{French Broadcast Programs Corpus}

%In order to build an ASR and a  PPSl. We use the data from the diffrent French projects:
The data used in our protocol comes from different broadcast collections in French:

\begin{itemize}
\item  Subset of  \textit{Quaero}\footnote{http://www.quaero.org} data which contains 41h of broadcast speech from different French radio and television programs on various subjects.

\item Data from \textit{ETAPE} \cite{gravier2012etape} project which includes 37h of radio and television programs (mainly spontaneous speech with overlapping speakers).

\item Data from \textit{ESTER 1 \& ESTER 2} \cite{galliano2005ester} containing 111h of transcribed audio, mainly from French and African radio programs (mix of prepared and more spontaneous speech: anchor speech, interviews, reports). %The main objective is to produce lexical transcripts, in order to extract high-level semantic information.  

\item Data from \textit{REPERE}  \cite{kahn2012presentation}: 54 hours of transcribed shows (spontaneous, such as debates) and TV news.  %Initially, REPERE was dedicated to multimodal person recognition in French TV shows.

\end{itemize}

As described in Table 1, the full data contains non spontaneous speech (NS) and spontaneous speech (S).
The data used to train our ASR system (\textit{Train$_{Acoustic}$}) is selected from the non-spontaneous speech style that corresponds mainly to broadcast news.
The data used for performance prediction (\textit{Train$_{pred}$} and \textit{Test$_{pred} $}) is a mix of both speech styles (S and NS). It is important to mention that shows in \textit{Test$_{Pred}$} data set were unseen in the \textit{Train$_{Pred}$} and vice versa. Moreover, more challenging (high WERs) shows were selected for  \textit{Test$_{Pred}$}.

 \begin{table}[thb]

  \centering
  \begin{tabular}{llrr}
    \toprule
        & Train$_{Acoustic}$ & Train$_{Pred }$  & Test$_{Pred} $ \\
    \midrule
    
    NS  & 100h51 &  30h27  & 04h17\\
   
    S    & - & 59h25 & 04h42 \\ 
    \hline
   Duration  & 100h51 & 89h52 &08h59 \\
   \textbf{WER}  & - & \textbf{22.29} &\textbf{ 31.20} \\
    \bottomrule
  \end{tabular}
   \label{tab:data}
    \caption{Distribution of our data set between non spontaneous (NS) and spontaneous (S) styles}
 \end{table}

%  \begin{table}[thb]
% 
%  \centering
%  \begin{tabular}{llrl}
%    \toprule
%         Source   & Show & Words  & WER \\
%    \midrule
%   
%    \multicolumn{4}{c}{\textit{Non Spontaneous (NS)} } 
%      \\ \hline
%    Quaero   &Arte News (AN) &  3714 &  12.21\\
%    ESTER 2 &Tvme (T) & 10698 & 18.44\\
%    Quaero  &FR Culture TEMPS (FCT) &   10092 & 20.92\\
%    Quaero   & Fab Histoire (FH) & 10018 & 22.76\\
%    ESTER 2  &Africa1 (A1)  &   15249 & 25.41\\ 
%    \hline
%      \multicolumn{4}{c}{\textit{Spontaneous (S)} } \\
%    \hline
%	Quaero & Ce Soir Ou Jamais (CSOJ)&   10993 &  28.74\\
%	REPERE &Planete Showbiz (PS)& 15943 & 36.74\\
%	REPERE &Culture Et Vous (CV)&  16028 & 39.79\\
%	ETAPE & La Place Du Village (PV)&   20397 &  45.15\\
%    \bottomrule
%  \end{tabular}
%   \caption{Performance on Test$_{Pred}$ dataset (WER) }
%  \label{tab:perfdevpred}
% \end{table}
 
%Table~\ref{tab:perfdevpred} displays the true performance (WER) on Test$_{Pred}$ dataset. 
Our shows with spontaneous speech logically have a higher WER (from 28.74\% to 45.15\% according to the program) compared to the shows with non-spontaneous speech (from 12.21\% to 25.41\% according to the broadcast program)\footnote{Detailed results per broadcast programs not shown here due to space constraints}. This S/NS division will allow us to compare our performance prediction systems on different types of documents with \textit{non spontaneous} and \textit{spontaneous} speech. 
% it means that these shows are difficult to transcribe by the system ASR. We wish to evaluate the quality of the predictions models in these two styles of speech.
\subsection{ASR system used}
To obtain speech transcripts (ASR outputs) for the prediction model, we built our own French ASR system based on the KALDI toolkit \cite{povey2011kaldi} (following a standard Kaldi recipe). A hybrid HMM-DNN system was trained using Train$_{Acoustic}$  (100 hours of broadcast news from ESTER, REPERE, ETAPE and Quaero). A 5-gram language model was trained from several French corpora (3323M words in total - from EUbookshop, TED2013, Wit3, GlobalVoices, Gigaword, Europarl-v7, MultiUN, OpenSubtitles2016, DGT, News Commentary, News WMT, LeMonde, Trames, Wikipedia and transcriptions of our \textit{Train$_{Acoustic}$} dataset) using SRILM toolkit \cite{stolcke2002srilm}. For the pronunciation model, we used lexical resource BDLEX \cite{de1998bdlex} as well as  automatic grapheme-to-phoneme (G2P)\footnote{\url{http://lia.univ-avignon.fr/chercheurs/bechet/download/lia_phon.v1.2.jul06.tar.gz}} transcription to find pronunciation variants of our vocabulary (limited to 80k). 

%\textbf{Description of the acoustic model} \\
%For the acoustic modeling, we followed standard Kaldi recipes.  The acoustic features are 13 dimensions Mel-Frequency Cepstral Coefficients (MFCC) 
%with Linear Discriminative Analysis (LDA) and Maximum Likelihood Linear Transform (MLLT) applied to 7-splice (3 left and 3 right context) frames
%and projected to a 40-dimensions space. These features are used to train a conventional triphone GMM acoustic model.  
%Then a speaker dependent model is trained by applying a feature-based Maximum Likelihood Linear Regression (fMLLR) to the acoustic features. 
%The last model is a hybrid HMM-DNN where the DNN is trained to map fMLLR transformed features to the corresponding HMM tied states.
%All GMM models have 40k gaussians and the DNN has 4 hidden layers (of dimension 1024).
% IL ME FAUT DES PRECISIONS sur : nb de gaussiennes, nombre de couches, largeur du DNN, nb d'epochs pour le training 

%All GMM models contains about 150k gaussians and the DNN contains 5 hidden layers

\subsection{Evaluation}
The LNE-Tools~\cite{conf/interspeech/Galibert13a} are used to evaluate the ASR performance. Overlapped speech and empty utterances are removed. We obtain 22.29\% WER on Train$_{pred}$ and 31.20\% on Test$_{pred}$ (see Table 1).
In order to evaluate \textit{WER} prediction task,  we use \textit{Mean Absolute Error} (\textit{MAE}) metric defined as: 
\begin{equation}
  MAE = \frac{\sum_{i=1}^{N}|WER_{Ref}^i-WER_{Pred}^i|}{N}
  \label{MAE}
\end{equation}

where \textit{N} is the number of units (utterances or files).

%INTERSPEECH In the case of the classification according to the CLASS categories defined in Table~\ref{tab:dist4class}, we use a standard accuracy score (number of correct predictions divided by total number of units). 
We also use Kendall's rank correlation coefficient $\tau$ 
%and Spearman ($\rho$) 
between real and predicted WERs, at the utterance level.

% 
%    \textbf{Accuracy}: This metric is used to evaluate the classification task, it compares a class pair (class$_{ref}$, class$_{pred}$) in order to calculate the total number of correct predictions  and divide it on the total number of utterances. Accuracy is defined as follows:
%     \begin{equation}
%  Accuracy = \frac{Number\ of\ correct\ predictions}{Number\ of\ utterances}  
%  \label{Accuracy}
%\end{equation}
 
%    \textbf{Rank correlation}
%unsing Kendall’s  $\tau$ correlation 
% 
%Spearman  $\rho$
% 

\section{ ASR performance prediction}
\label{sapp}
\subsection{Regression Baseline}

An open-source tool for automatic speech recognition quality estimation, \textit{TranscRater} \cite{jalalvand2016transcrater}, is used for the baseline regression approach.
%It was pro to perform ASR evaluation by passing the need of reference, transcripts and confidence information.
It requires \textit{engineered features} to predict the WER performance. These features are extracted for each utterance and are of several types: 
\textit{Part-of-speech (POS)} features  capture the plausibility of the transcription
from a syntactic point of view\footnote{\textit{Treetagger} \cite{schmid1995treetagger} is used for POS extraction in this study};
\textit{Language model (LM)} features capture the plausibility of the transcription according to a N-gram model (fluency)\footnote{We train a 5-gram LM on 3323M words text already mentioned};
\textit{Lexicon-based (LEX)} features are extracted from the ASR lexicon\footnote{A feature vector containing the frequency of phoneme categories in its prononciation is defined for each input word};
% (fricatives, liquids,  nasals consonants, nasals vowels, stops and  vowels in its pronunciation).
%A word may have several pronunciations but we only use the first one in our dictionary for our experiments.% to assign the features vector for each word.
% average frequency vectors assigned to each pronunciation variant in order to obtain a single representation for each word. 
\textit{Signal (SIG)} features capture the difficulty of transcribing the input signal (general recording conditions, speaker-specific accents)\footnote{For feature extraction, \textit{TranscRater} computes 13 MFCC (using  \textit{Opensmile}\cite{Eyben:2010:OMV:1873951.1874246}), their delta, acceleration and log-energy, F0, voicing probability, loudness contours and pitch for each frame. The SIG feature vector for the entire input signal is obtained by averaging the values of each frame}. 
This approach, based on \textit{engineered features}, can be considered as our baseline. One drawback is that its application to new languages requires to find adequate resources, dictionaries and tools which makes the prediction method less flexible. The next sub-section proposes a new (resource-free) prediction approach based on convolutional neural networks (CNNs) where features are learnt during training.

\subsection{Convolutional Neural Networks (CNNs)}
%The Convolutional Neural Networks (CNN) are very well known in computer vision. They were successful for image classification and captioning tasks. More 

%Recently, CNNs have shown interesting results in natural language processing when used for classification tasks (texts/documents). 
% In order to predict the four classes proposed in section \ref{sub:Overview}, we built upon the \textit{CNN-sentence} tool \cite{kim2014convolutional}. Its architecture is inspired from \cite{collobert2011natural}:
For WER prediction, we built our model using both \textit{Keras} \cite{chollet2015keras} and \textit{Tensorflow\footnote{https://www.tensorflow.org}}. 

\begin{figure}[htb]
\begin{minipage}[b]{\linewidth}
  \centering
  \centerline{\includegraphics[width=\linewidth]{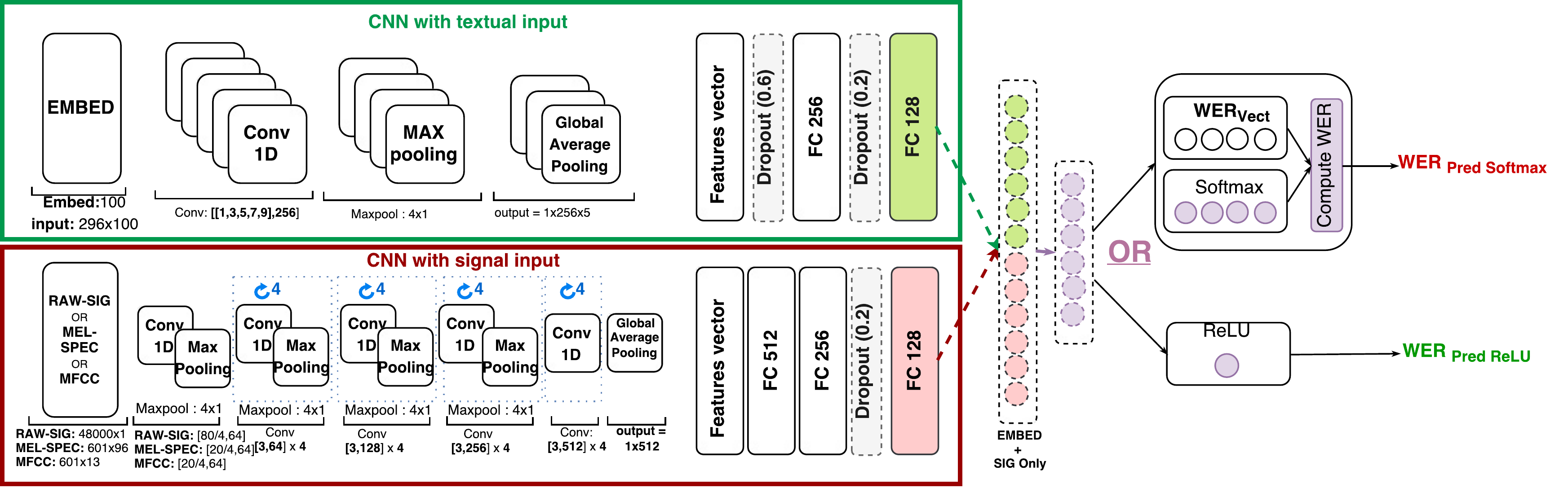}}
\end{minipage}

\caption{Architecture of our CNN with text (green) and signal (red) inputs - dotted lines correspond to  joint text+signal case}
\label{fig:architecture}
\end{figure}

\textbf{For pure textual input}, we propose an architecture inspired from \cite{kim2014convolutional} (green in Figure~\ref{fig:architecture}).
Input is an utterance padded to \textit{N} words (N is set as the length of the longest sentence in our full corpus) presented as a matrix \textit{EMBED} of size NxM (M is embedding size - our embeddings are obtained with Word2Vec \cite{mikolov2013}). 
% tool. These embeddings were trained on a big quantity of text (corpora used for ASR language model). The matrix is the concatenation of the vector representations of the words of a utterance. 
The convolution operation involves a filter $w$ which is applied to a segment of $h$ words to produce a new feature. For example, feature $c_i$ is generated from the words $x_{i:i+h-1}$ as:
%(Equation \ref{equ:featcnn}): 
\begin{equation}
c_{i} = f(w .x_{i:i+h-1}+b) 
  \label{equ:featcnn}
\end{equation}
 
Where \textit{b} is a bias term and \textit{f} is a non-linear function. This filter is applied to each word segment in the utterance to produce a \textit{feature map} \textit{c = [c$_{1}$, c$_{2}$…c$_{n-h+1}$]}. \textit{Max-pooling}~\cite{collobert2011natural} then takes the 4 largest values of $c$, which are then averaged.  $W$ filters provide a W-sized input to two fully-connected hidden layers (256 and 128)  followed respectively by dropout regularization (0.2 and 0.6) before WER prediction.

\textbf{For signal input}, we use the best architecture  ({\em m18})  proposed in \cite{CNNrawWav} (colored in red in Figure~\ref{fig:architecture}). This is a  deep CNN with 17 conv+max-pooling layers followed by global average pooling and  three hidden layers (512, 256 and 128 dimensions). A dropout regularization of 0.2 is added between the last two layers (256 and 128). We investigate several inputs to the CNN using \textit{Librosa} \cite{mcfee2015librosa}: raw signal, mel-spectrogram or MFCCs.

In order to predict the WER using CNN, we propose two different approaches:
\begin{itemize}
\item \textbf{CNN$_{Softmax}$}: 
we use $Softmax$ probabilities and an external fixed WER$_{Vector}$ to compute WER$_{Pred}$.  WER$_{Vector}$  and \textit{Softmax} output must have the same dimension. WER$_{Pred}$ is then defined as (expectation):
 % in order to have a probability distribution that we convert it into WER using the vector 
\begin{equation} WER_{Pred} = \sum_{C=1}^{NC} P_{Softmax}(C) * WER_{Vector} (C)
\label{cnnwerpred}\end{equation}
% INTERSPEECH With \textit{NC}  total number of classes and WER$_{center}$ assigned for each class as: \textit{VGQ}: WER$_{center}$= 5\%, \textit{GQ}: WER$_{center}$ = 20\%, \textit{BG}: WER$_{center}$= 40\% and \textit{VBQ}: WER$_{center}$= 125\%.
 With \textit{NC} as total number of classes.  In our experiment, we use 6 classes with WER$_{Vector}$=[0\%, 25\%, 50\%, 75\%, 100\%, 150\%],
  
\item \textbf{CNN$_{ReLU}$}: after the last FC layer, a \textit{ReLU} function returning a float value between 0 and  $+\infty$ estimates directly WER$_{Pred}$. 
\end{itemize}

% ZIED : Since \textit{CNN-Sentence} predicts classes, our purpose is to convert the classes obtained in WER by taking advantage of the Softmax probabilities assigned for each category. A WER$_{center}$ was assigned for each class, \textit{VGQ}: WER$_{center}$= 5\%, \textit{GQ}: WER$_{center}$ = 20\%, \textit{BG}: WER$_{center}$= 40\% and \textit{VBQ}: WER$_{center}$= 125\%.

\textbf{For joint use of both speech and text}, %we propose the architecture below (DECRIRE EN QUELQUES PHRASES)...
we merge the last hidden layers of both CNN \textit{EMBED} and CNN \textit{RAW-SIG (or MEL-SPEC or MFCC)} by concatenating and passing them through a new hidden layer before CNN$_{Softmax}$ or CNN$_{ReLU}$ (see dotted lines in the Figure~\ref{fig:architecture}) and we train the full network similarly.

% Laurent : While previously described regression approach predicts WER, CNN predicts performance category of table 1 (CLASS).Our CNN implementation is based on \textit{CNN-Sentence} \cite{kim2014convolutional}. The system input is made up of word embeddings (\textbf{explain!}). The architecture has one convolution layer followed by a max pooling layer. We use Dropout \cite{hinton2012improving} for regularization of the model. Finally, a softmax layer transforms and normalizes the output of the network into a probability distribution on the different possible performance categories.
%Consequently, in order to estimate predicted (EXACT) WER, we use softmax probabilities as following:

%\begin{equation}
%  WER_{pred} = \sum_{class=1}^{NC} (P_{softmax}(class) * WER_{center})
%  \label{cnnwerpred}
%\end{equation}

%With \textit{NC} is the total number of classes and WER$_{center}$ is assigned for each class as: \textit{VGQ}: WER$_{center}$= 5\%, \textit{GQ}: WER$_{center}$ = 20\%, \textit{BG}: WER$_{center}$= 40\% and \textit{VBQ}: WER$_{center}$= 125\%.

%Unlike the expensive  \textit{engineered features}, CNN features are trained from word embeddings. These characteristics are modified by the neural network until the desired behavior is obtained.
\vspace{-0.5cm}

\section{Experiments and Results}
\label{sear}
In this section,  \textit{Regression} and \textit{CNN} approaches are compared for  ASR performance prediction.
The \textit{Regression}  uses several engineered features extracted from the ASR output (POS, LEX, LM, SIG)  while the \textit{CNN} is based on features learnt from the ASR output and from the signal only.   %EMBED+RAW-SIG.    
For the CNN, we randomly select 10\% of the Train$_{Pred}$ data as a \textit{Dev} set. 10 different model trainings (cross-validation) with 50 epochs are performed.
% in case  of a single input and 50 epochs in case of multiple inputs . 
Training is done with the $Adadelta$ update rule \cite{adelta} over shuffled mini-batches. We use $MAE$ as both loss function and evaluation metric.    After training, we take the model (among 10) that led to the best \textit{MAE} on \textit{Dev} set and report its performance on Test$_{Pred}$. 
We investigate several inputs to the CNN: %For each input type, we use specific parameters:  
\begin{itemize}
\item Textual  (ASR transcripts) only  (\textbf{EMBED}): input matrix has dimension 296x100 (296 is length of longest ASR hypothesis in our corpus ; 100 is dimension of word embeddings pre-trained on our large text corpus of 3.3G words)\footnote{We use filter window sizes \textit{h} of [1, 3, 5, 7, 9] with 256 filters per size},
 
\item Raw signal only  (\textbf{RAW-SIG}): models are trained on six-second speech turns and sampled at  8khz (to avoid memory issues). Short speech turns ($<6s$) are padded with zeros. Our input has dimension 48000 x 1\footnote{The detailed parameters of the filters are given in Figure~\ref{fig:architecture}},

\item Spectrogram only  (\textbf{MEL-SPEC}): we use same configuration as for raw signal ; we have 96-dimensional vectors (each dimension corresponds to a particular mel-frequency range) extracted every 10ms (analysis window is 25ms). Our input has a dimension 601x96\footnote{The detailed parameters of the filters are given in Figure~\ref{fig:architecture}},
\item \textbf{MFCC} features only: we compute 13 MFCCs every 10ms to provide the CNN network with an input of dimension 601x13,

\item Joint (textual and signal) inputs (\textbf{EMBED}+\textbf{RAW-SIG}\footnote{or EMBED+MEL-SPEC or EMBED+MFCC}): in that case, we concatenate last hidden layers of both textual and signal inputs (dotted lines of Figure~\ref{fig:architecture}).
\end{itemize} 
\vspace{-0.5cm}
\subsection{Regression (baseline) and CNN performances}

%\begin{table}[th!]
%\centering \begin{tabular}{ lrrrr  }
%\hline Model & POS & LEX & LM & SIG \\\midrule
%\textbf{Regression} & 25.95 & 25.78 &\textbf{ 24.19} & 25.86 \\\hline
%\end{tabular}
%\caption{Regression model  \cite{jalalvand2016transcrater} evaluated at utterance level with MAE for each feature on Test$_{pred}$}
%\label{tab:REGALLFEAT}
%\end{table}
%% \color{blue}
%
%OG Recheck once all the results are in, it's a little redundant and the tables are not explicitly referred to.

%Table \ref{tab:REGALLFEAT} presents the baseline (regression) performance using features POS, LEX, LM and SIG separately. 
The lines \textit{Regression} of table 2 show results obtained with combined features\footnote{MAE using single POS, LEX, LM and SIG features is respectively 25.95\%, 25.78\%, 24.19\% and 25.86\% }.
We can observe that  the best performance is obtained with POS+LEX+LM features (MAE of 22.01\%) while adding the SIG does not really improve the model (MAE of 21.99\%).  This inefficiency of SIG features in regression models was also observed in  \cite{jalalvand2016transcrater}.

\begin{table}[th]
\centering
\begin{tabular}{ |l|l|l|c|}  
\hline  Model & Input &   $MAE$  &  $\tau$ 

%%%%%%%%%%%%%%%% Textual features
\\\hline \multicolumn{4}{|c|}{\cellcolor{lightgray!50} \textbf{Textual features}}         
\\ \hline    \textbf{Regression} &POS+LEX+LM& 22.01  & \textbf{44.16} 
\\ \hline\hline   \textbf{CNN$_{Softmax}$} &EMBED& \textbf{21.48} & 38.91
\\ \hline     \textbf{CNN$_{ReLU}$} &EMBED& 22.30  & 38.13
\\ \hline    \multicolumn{4}{|c|}{\cellcolor{lightgray!50} \textbf{Signal features}} 

%%%%%%%%%%%%%%%%SIG Only

\\ \hline   \textbf{Regression} & SIG& 25.86  & 23.36

\\ \hline\hline    \textbf{CNN$_{Softmax}$} & RAW-SIG   & 25.97 & 23.61
\\ \hline   \textbf{CNN$_{ReLU}$} & RAW-SIG   & 26.90 & 21.26
         
\\ \hline\hline  \textbf{CNN$_{Softmax}$} & MEL-SPEC   & 29.11  & 19.76 
\\ \hline   \textbf{CNN$_{ReLU}$} & MEL-SPEC   & 26.07 & 24.29 
 
\\ \hline\hline    \textbf{CNN$_{Softmax}$} & MFCC  &\textbf{25.52}  & \textbf{26.63}  
\\ \hline   \textbf{CNN$_{ReLU}$} & MFCC & 26.17  & 25.41
%%%%%%%%%%%%%%%% Textual and Signal features

\\ \hline  \multicolumn{4}{|c|}{\cellcolor{lightgray!50} \textbf{Textual and Signal features}} 
\\ \hline    \textbf{Regression} & POS+LEX+LM+SIG & 21.99  & 45.82
\\ \hline  \hline  \textbf{CNN$_{Softmax}$} &  EMBED+RAW-SIG   & \textbf{19.24}  & \textbf{46.83}
\\ \hline    \textbf{CNN$_{ReLU}$} &  EMBED+RAW-SIG    & 20.56  & 45.01
         
\\ \hline\hline    \textbf{CNN$_{Softmax}$} & EMBED+MEL-SPEC   & 20.93  & 40.96
\\ \hline  \textbf{CNN$_{ReLU}$} & EMBED+MEL-SPEC   & 20.93  & 44.38
 
\\ \hline\hline      \textbf{CNN$_{Softmax}$} & EMBED+MFCC  & 19.97 & 44.71
\\ \hline     \textbf{CNN$_{ReLU}$} & EMBED+MFCC & 20.32  & 45.52 
\\ \hline
\end{tabular}
\label{tab:transcRaterVScnn}
\caption{Regression \textit{vs} CNN$_{Softmax}$ \textit{vs} CNN$_{ReLU}$ evaluated at utterance level with MAE or $\tau$ on Test$_{pred}$}  
\end{table}

 \vspace{-0.2cm}
As for the use of textual features only, CNN$_{Softmax}$ and CNN$_{ReLU}$ are equivalent (better MAE but lower $\tau$) than the regression model that uses engineered features. CNN$_{Softmax}$ shows better performance than CNN$_{ReLU}$. 
%However, in terms of rank-correlation Kendall's $\tau$ and Spearman's $\rho$ , the regression model is better than both CNNs.
Concerning signal features only, ASR performance prediction is a difficult task with all MAE above 25\%. However, among the different signal inputs to the CNN, simple MFCCs lead to the best performance both for MAE and $\tau$.
While the joint use of textual and signal features did not work for the regression baseline, the combination of inputs for CNNs lead to improved results. The best performance is obtained with CNN$_{Softmax}$ (EMBED + RAW-SIG) which outperforms a strong regression baseline (MAE is reduced from 21.99\% to 19.24\%, while $\tau$ is improved from 45.82\% to 46.83\%),  \textit{Wilcoxon Signed-rank Test}\footnote{\url{http://www.r-tutor.com/elementary-statistics/non-parametric-methods/wilcoxon-signed-rank-test}} confirms that the difference is significant with p-value of 4e-08.

\subsection{Analysis of predicted WERs}

Table~\ref{tab:wer_pred} shows the predicted WERs (at collection level) for both regression and (best) CNN approaches for different speaking styles (spontaneous and non spontaneous). Overall, the predicted WER on non spontaneous (NS) and  spontaneous (S) speech is very good for the \textit{CNN} approach. WER$_{Pred}$ is at -2.54\% on non-spontaneous speech and at -4.84\% on spontaneous speech. On the other hand, while efficient on non-spontaneous speech, regression fails to predict performance (-10,11\%) on spontaneous speech. 

\begin{table}[th]
\centering
\begin{tabular}{ lrrr  }
\hline
& NS & S  & NS + S   \\\hline
WER$_{REF}$ & 21.47 &  38.83 & 31.20 \\\hline 
WER$_{Pred }$ Regression &   \textbf{22.08}  & 28.72 &  25.82  \\
WER$_{Pred} $ CNN$_{Softmax}$  &   18.93  & \textbf{33.99}  & \textbf{27.37} \\
%WER$_{Pred} $ CNN$_{ReLU}$  &   \textbf{-}  & \textbf{-} & \textbf{-}  \\
\hline \#Utterances  & 3,1k & 3,7k & 6,8k  \\
\#Words$_{REF}$ & 49.8k & 63.3k&  113,1k \\ 
\hline
\end{tabular}
\caption{Regression \textit{vs} CNN$_{Softmax}$ predicted WERs (averaged over all utt.) per speaking style (NS/S) on Test$_{pred}$ } 
\label{tab:wer_pred}
\end{table}

\begin{figure}[htb]
\begin{minipage}[b]{.3\linewidth}
  \centering
  \centerline{\includegraphics[width=3.1cm]{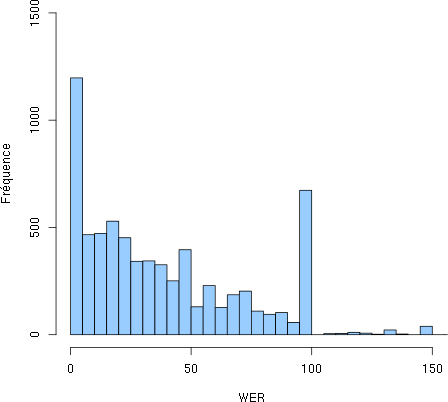}}
%  \vspace{1.5cm}
  \centerline{(a) REF}\medskip %TranscRater  All Features$_{Feat}$}\medskip
\end{minipage}
\begin{minipage}[b]{.3\linewidth}
  \centering
  \centerline{\includegraphics[width=3.1cm]{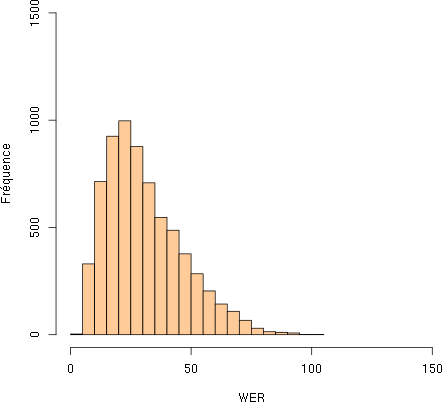}}
%  \vspace{1.5cm}
  \centerline{(c) Best REG }\medskip %TranscRater  All Features$_{Feat}$}\medskip
\end{minipage}
\begin{minipage}[b]{.3\linewidth}
  \centering
  \centerline{\includegraphics[width=3.1cm]{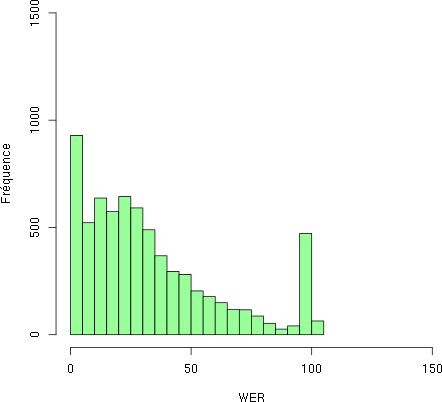}}
%  \vspace{1.5cm}
  \centerline{(b) Best CNN}\medskip % Softmax All$_{Feat}$}\medskip
\end{minipage}
\vspace{-0.3cm}
\caption{Distribution of speech turns according to their WER: (a) real (b) predicted by regression (c)  predicted by CNN}
\label{fig:distrib}

\end{figure}

Figure~\ref{fig:distrib} analyzes WER prediction at utterance level\footnote{Model outputs available on \url{http://www.lne.fr/LNE-LIG-WER-Prediction-Corpus}} . It shows the distribution of speech turns according to their real or predicted WER. It is clear that CNN prediction allows to approximate the true WER distribution on Test$_{pred}$ while regression seems to build a gaussian distribution around the mean WER observed on training data. It is also remarkable that the two  peaks  at WER=0\% and WER=100\% can be predicted correctly by our CNN model.

\vspace{-0.6cm}

%%%%%%%%%%%%%% INTERSPEECH %%%%%%%%%%%%%%%%
%On the contrary, CNN gets closer to the true WER on spontaneous speech while it overestimates its value on non spontaneous shows. This result confirms that the \textit{Regression} and \textit{CNN} approaches might be complementary. It is also important to note that there is a large room for improvement with the CNN model: optimizing the dimension of the softmax layer, optimizing CNN architecture and, as a contrast, adding POS and LM information into CNN itself to allow a fair comparison with the \textit{Regression} model.
%%%%%%%%%%%%%%%%%%%%%%%%%%%%%%%%%%%%%%%%%%%%

%overall predicted WER  well on non spontaneous speech  that the approach of regression has a good perfomance compared with CNN model on the non-spontaneous data, with a WER-pred approximately equal to the WER$_{REF}$. On the other hand, the CNN model predicts a WER 4 points lower than the WER$_{REF}$.
%On the spontaneous data, the CNN model shows a better prediction then the regression in term of WER, with a 5\% difference between the predicted WER and the WER$_{REF}$, however, the difference between the predicted WER and the WER$_{REF}$ is 10\% for the approach of regression. On the whole show, the CNN model is better than the regression, it has a WER$_{CNN}$=30.08 wgucg us very close to the WER$_{REF}$, however, the regression model is at -6\% of the WER$_{REF}$.
%%%%%%%%%%%%%%%%%%%%%%%%%%%%%%%%%%%%%%%%%%%%%%%%%%%%%%%%%%%%%%%%%%%%%%%%%

%%%%%%%%%%%%%%%%%%%%%%%

\section{Conclusions}
\label{sc}
 \vspace{-0.2cm}
This paper presented an evaluation framework for evaluating ASR performance prediction on unseen broadcast programs.
% (French corpus, evaluation metrics). We contrasted our results on two different speech styles (non spontaneous and spontaneous). We focused on two different performance prediction approaches: one based on regression (\textit{WER} prediction task) and another based on convolutional neural networks. 
CNNs were very efficient encoding both textual (ASR transcript) and signal to predict WER. 
%Since CNNs are specialized in predicting categories / classes, we presented two new ways (CNN $_{Softmax}$ and $_{ReLU}$) to convert their outputs into a continuous \textit{WER} value. Our results show that CNN $_{Softmax}$ is a little better than CNN $_{ReLU}$. 
%Using  textual and raw signal only,CNN approach showed a good performance level  at \textit{MAE} and $\tau$  compared to CNN MEL-SPEC, CNN MFCC, and the regression approach which use textual and signal features. 
%However, looking at the \textit{WER} prediction at a broader grain (collection of spontaneous and non spontaneous speech shows), the CNN model  was better at predicting the overall WER on spontaneous data (S) and on the whole collection (NS+S). Regression approaches had difficulties in predicting the ASR performance on spontaneous speech. 
%Our experiments showed that \textit{MAE} is not the relevant metric for the \textit{WER} prediction task.
Future work will be dedicated to the analysis of signal and text embeddings learnt by the CNN and their relation to conditioning factors such as speech style, dialect or noise level.
 \newpage
\newlength{\bibitemsep}\setlength{\bibitemsep}{.2\baselineskip plus .05\baselineskip minus .05\baselineskip}
\newlength{\bibparskip}\setlength{\bibparskip}{0.4pt}
\let\oldthebibliography\thebibliography
\renewcommand\thebibliography[1]{%
  \oldthebibliography{#1}%
  \setlength{\parskip}{\bibitemsep}%
  \setlength{\itemsep}{\bibparskip}%
}

\bibliographystyle{IEEEbib}
 
\bibliography{icassp2018-prediction}

\end{document}